\PassOptionsToPackage{table,dvipsnames}{xcolor}
\documentclass[10pt,twocolumn,letterpaper]{article}

\pdfoutput=1

\usepackage{iccv}

\definecolor{iccvblue}{rgb}{0.21,0.49,0.74}

\usepackage{multirow}

\usepackage{algorithm}
\usepackage{algorithmic}
\usepackage{amsmath}

\usepackage[pagebackref=true,breaklinks=true,colorlinks,bookmarks=false]{hyperref}

\definecolor{myred}{HTML}{B22222}   
\definecolor{mygreen}{HTML}{006400} 

\usepackage{adjustbox} 

\definecolor{Orange}{HTML}{F4782F}
\definecolor{DeepBlue}{HTML}{0082C9}
\definecolor{MyPink}{HTML}{F1708E}
\definecolor{figure3b}{HTML}{173047}
\definecolor{Pipeline1}{HTML}{EB6C8B}
\definecolor{Pipeline2}{HTML}{012F45}
\definecolor{Pipeline3}{HTML}{EE761C}

\usepackage{booktabs}
\newcommand{\Xhline}[1]{\noalign{\hrule height #1}}
\usepackage{fontawesome}

\definecolor{yzybest}{rgb}{0.98, 0.8, 0.8} 
\definecolor{yzysecond}{rgb}{0.99, 0.88, 0.77} 
\definecolor{yzythird}{rgb}{1.0, 1.0, 0.8} 

\newcommand{\bestcolor}{\cellcolor{yzybest}}
\newcommand{\secondcolor}{\cellcolor{yzysecond}}

\title{\vspace{-1.1cm}X-Field: A Physically Grounded Representation for 3D X-ray Reconstruction}
\author{Feiran Wang,\negmedspace\textsuperscript{1,\textasteriskcentered}
Jiachen Tao,\negmedspace\textsuperscript{2,\textasteriskcentered}
Junyi Wu,\negmedspace\textsuperscript{2,\textasteriskcentered}
Haoxuan Wang,\negmedspace\textsuperscript{2}
Bin Duan,\negmedspace\textsuperscript{3}
\\
Kai Wang,\negmedspace\textsuperscript{4}
Zongxin Yang,\negmedspace\textsuperscript{5}
Yan Yan\textsuperscript{2,$\dagger$} \vspace{1ex} \\
\textsuperscript{1}Illinois Institute of Technology \qquad \textsuperscript{2}University of Illinois Chicago \qquad \textsuperscript{3}University of Michigan \qquad \\
\textsuperscript{4}National University of Singapore \qquad \textsuperscript{5}Harvard University
}

\begin{document}

\twocolumn[{
	\maketitle
    \vspace{-2em}
	\centerline{\hspace{-2mm}
		\includegraphics[width=1\linewidth,trim={4pt 4pt 4pt 4pt}]{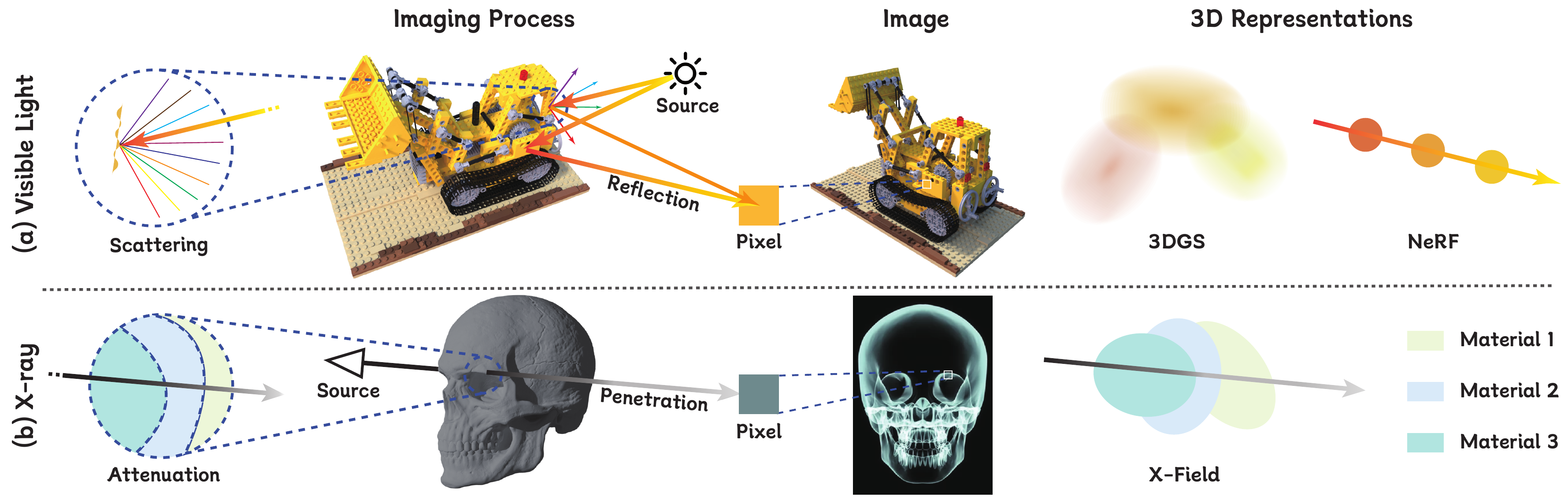}}
	\vspace{-1mm}
    \captionsetup{hypcap=false}
	\captionof{figure}
	{
		\textbf{Comparison of Imaging Processes and Corresponding 3D Representations.} (a) Visible light interacts with surface mainly through scattering and reflection. NeRF~\cite{nerf} and 3DGS~\cite{3dgs} model this process by accumulating directional light rays. (b) Rooted in X-rays' attenuation and penetration properties, our X-Field models the radiological density of different materials to reveal internal structure.
	}
	\vspace{4mm}
    \vspace{-0.2em}
	\label{fig:teaser}
}]

\begingroup
\renewcommand{\thefootnote}{}
\footnotetext[1]{Project Page: \url{ https://brack-wang.github.io/XField/}}
\footnotetext[2]{$^*$Equal Contribution{  } $^\dagger$Corresponding Author}
\endgroup

\setcounter{footnote}{10}

\begin{abstract}
\vspace{-10mm}

X-ray imaging is indispensable in medical diagnostics, yet its use is tightly regulated due to potential health risks. To mitigate radiation exposure, recent research focuses on generating novel views from sparse inputs and reconstructing Computed Tomography (CT) volumes, borrowing representations from the 3D reconstruction area.
However, these representations originally target visible light imaging that emphasizes reflection and scattering effects, while neglecting penetration and attenuation properties of X-ray imaging.
In this paper, we introduce \textbf{X-Field}, the first 3D representation specifically designed for X-ray imaging, rooted in the energy absorption rates across different materials. 
To accurately model diverse materials within internal structures, we employ 3D ellipsoids with distinct attenuation coefficients. To estimate each material's energy absorption of X-rays, we devise an efficient path partitioning algorithm accounting for complex ellipsoid intersections.
We further propose hybrid progressive initialization to refine the geometric accuracy of X-Filed and incorporate material-based optimization to enhance model fitting along material boundaries.
Experiments show that X-Field achieves superior visual fidelity on both real-world human organ and synthetic object datasets, outperforming state-of-the-art methods in X-ray Novel View Synthesis and CT Reconstruction.

\end{abstract}
\vspace{-5mm}
\section{Introduction}
\label{sec:intro} 
\vspace{-1mm}


X-ray imaging is a cornerstone of Computed Tomography (CT) reconstruction~\cite{feldkamp1984practical, andersen1984simultaneous, hounsfield1973computerized, cormack1963representation}, providing critical insights into internal structures for clinical diagnostics.
X-rays undergo progressive attenuation while penetrating materials until the residual energy reaches the detector, forming an X-ray projection.
Traditional CT reconstruction relies on hundreds of X-ray projections pulsating from various angles to recover a density field~\cite{brenner2007computed, smith2009radiation, hricak2011managing}.
However, acquiring such a vast number of projections exposes patients to high doses of ionizing radiation, posing substantial health risks.
Consequently, X-ray reconstruction has gained increasing attention~\cite{chen2022tensorf, zha2022naf, sax_nerf, x_gaussian, r2_gaussian, ruckert2022neat}, aiming to synthesize novel X-ray views from a sparse set of input projections and facilitate the reconstruction of high-quality CT volumes.

Recent advances in 3D reconstruction from sparse inputs~\cite{3dgs, nerf, mueller2022instant} have laid the groundwork for X-ray reconstruction~\cite{sax_nerf, x_gaussian, r2_gaussian}.
Yet, current 3D reconstruction techniques were originally designed for visible light imaging, as illustrated in Figure~\ref{fig:teaser}(a).
When visible light rays interact with a surface, despite minor absorption, most wavelengths are \textit{reflected} or \textit{scattered} into multiple directions \cite{born2013principles}.
As a result, light rays from different directions can accumulate to form a single pixel, producing varying appearances of the same surface depending on the viewpoint.
Neural Radiance Fields (NeRF)~\cite{nerf} employ deep networks to implicitly encode directional light information at each sampled point, conditioned by spatial positions and view directions to model appearance variation.
Similarly, 3D Gaussian Splatting (3DGS) \cite{3dgs} projects anisotropic ellipsoids onto the image plane 
and accumulates the projected Gaussian splats to determine the pixel colors.
Both methods reconstruct scenes considering the accumulation of multi-directional rays, thus well-suited for visible light imaging.

However, X-ray imaging operates on fundamentally different principles from visible light imaging, as depicted in Figure~\ref{fig:teaser}(b).
First, X-rays are a form of high-energy electromagnetic 
 radiation that is more capable of \textit{penetrating} objects~\cite{kak2001principles}, whereas visible light, emitted from natural sources, has lower energy and is primarily reflected by surfaces.
Second, X-rays progressively \textit{attenuate} as they pass through an object, traveling in a nearly straight line before reaching the detector on the opposite side.
Third, the pixel intensity is primarily determined by energy absorption from the corresponding X-ray, rather than by the combined contributions of visible light from multiple directions.

Therefore, X-ray imaging reveals the internal structure and material composition of a 3D object, rather than the surface color and geometry captured in visible light imaging. 
Inside an object, numerous materials with distinct densities are distributed throughout its structure.
As X-rays pass through these materials, each absorbs energy in proportion to its radiological density.
To faithfully adhere to the principles of X-ray imaging,  X-ray reconstruction demands a representation that accurately models the radiological density across various materials, which remains underexplored. 



In this paper, we present \textbf{X-Field}, a physically grounded ellipsoid representation specifically designed to model X-ray attenuation as it penetrates materials during imaging.
To differentiate various materials, X-Field represents an object's structure using ellipsoids with distinct attenuation coefficients.
Then, to track the distance an X-ray travels through each material, we derive an explicit form for segment length in the normalized device coordinate space.
For precise pixel intensity calculation, we express the attenuation integral along the ray path as a finite Riemann sum of the attenuation coefficients and segment lengths.
Yet, during optimization, ellipsoids with different attenuation coefficients may overlap.
To ensure that each position in 3D space corresponds to a single material, we further design a first-pass precedence strategy, resolving ellipsoid ambiguities in overlapping regions. Furthermore, we propose Material-Based Optimization, which adaptively refines ellipsoid placement by splitting them along material boundaries guided by the local attenuation gradient.


We evaluate X-Field on X-ray reconstruction across various modalities, including real-world human organ datasets and simulated object datasets.
Furthermore, we compare X-Field with existing methods in sparse-view CT reconstruction based on Novel View Synthesis (NVS) results.
Experiments demonstrate that X-Field outperforms state-of-the-art (SOTA) methods in both NVS and CT reconstruction.
Notably, our method achieves optimal results within 10 minutes using only 10 input X-ray projections, delivering substantial PSNR improvements of 2.44 dB for X-ray NVS and 3.98 dB for CT reconstruction over SOTA methods.

\begin{figure*}[t]
    \centering
    \includegraphics[width=\textwidth]{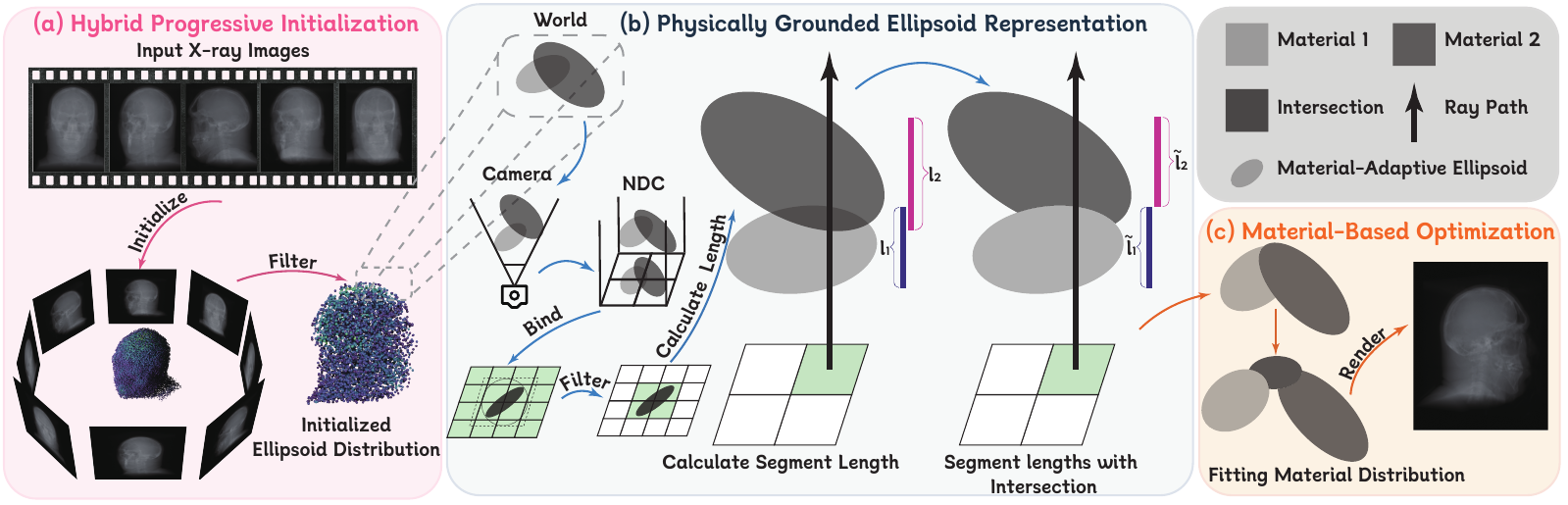}
    \captionsetup{type=figure}
    \vspace{-2.0em} 
    \caption{\textbf{Overview of X-Field.} \textbf{\textcolor{MyPink}{(a) Hybrid Progressive Initialization.}} We begin with X-ray images to construct a coarse initialization using combined iterative methods. 
    \textbf{\textcolor{figure3b}{(b) Physically Grounded Ellipsoid Representation.}} We transform initialized ellipsoids into NDC space and associate them with pixels. We then compute the attenuation integral along the ray, considering segment length and ellipsoid intersections. \textbf{\textcolor{Orange}{(c) Material-Based Optimization.}} Our optimization captures detailed materials' boundaries for high-quality rendering.}
    \vspace{-0.6em} 
    \label{fig:pipeline}
    \vspace{-4pt}
\end{figure*}

\section{Related Work}
\subsection{X-ray Novel View Synthesis}


3D X-ray reconstruction consists of two tasks: X-ray novel view synthesis and CT reconstruction \cite{sax_nerf}.
X-ray NVS \cite{sax_nerf, x_gaussian, zha2022naf, r2_gaussian} is essential for medical diagnostics, as it reconstructs images from sparse view inputs, significantly reducing patient exposure to X-ray radiation. Motivated by the fact that visible light and X-rays are both electromagnetic waves, two representation methods originally designed for visible light have been adopted for X-ray reconstruction: Neural Radiance Fields (NeRF)~\cite{nerf}, which employ multi-layer perceptrons for implicit encoding, and 3D Gaussian Splatting (3DGS)~\cite{3dgs}, which uses Gaussian ellipsoids for explicit representation. 
NeRF-based methods, such as NAF~\cite{zha2022naf}, improve efficiency by incorporating hash tables, while SAX-NeRF~\cite{sax_nerf} introduces a transformer architecture to better model 3D structural dependencies.
3DGS-based approaches, such as X-Gaussian\cite{x_gaussian}, replace the spherical harmonics color in 3DGS with a response function, addressing differences between visible light color and X-ray intensity. R$^2$-Gaussian\cite{r2_gaussian} further corrects integration bias for more accurate density retrieval.

However, neither representation has a comprehensive consideration of the attenuation and penetration properties of the X-ray. Instead, we introduce material-adaptive ellipsoids with distinct attenuation coefficients and consider the X-ray penetration length within the materials, ensuring more accurate and artifact-free reconstructions.


\subsection{Computed Tomography Reconstruction}
CT reconstruction is of vital importance for domains such as medical diagnosis \cite{cormack1963representation, cormack1964representation, elbakri2003segmentation, hounsfield1973computerized, hounsfield1980computed, sauer1993local}, biology \cite{kiljunen2015dental, lucic2005structural, donoghue2006synchrotron}, industrial inspection \cite{sun2012overview, dechiffre2014industrial}, and security screening \cite{von2010multi}. Early CT reconstruction works~\cite{kak2001principles} can be categorized into analytical approaches \cite{feldkamp1984practical, yu2006region} and iterative techniques \cite{andersen1984simultaneous, pan2006variable, sidky2008image}. Analytical methods such as FDK~\cite{feldkamp1984practical} perform a filtered back-projection of acquired projections. Iterative methods such as SART~\cite{andersen1984simultaneous} and CGLS~\cite{bjorck1979conjugate} reconstruct CT images by iteratively refining the solution through algebraic projection corrections and solving a least-squares optimization problem, respectively. However, these methods typically require hundreds to thousands of X-ray images, leading to increased radiation exposure.

Fortunately, X-ray NVS offers an effective alternative to CT reconstruction. By generating diverse novel views from sparse X-ray inputs~\cite{x_gaussian, r2_gaussian, sax_nerf, ruckert2022neat, li2023sparse}, it enables reconstruction of high-quality CT volumes from limited data. We further reduce the number of required X-ray images to as few as five while achieving superior performance both qualitatively and quantitatively compared to other baselines, highlighting its practicality for real-world applications.

%

\section{Method}

In this section, we introduce our Physically Grounded Ellipsoid Representation in Sec. \ref{4.1}, including (1) the formulation of X-ray Physical Field, (2) the Material-Adaptive Ellipsoids, (3) the algorithms for calculating segment lengths with intersections, and (4) the overlap filtering method.
Then, we propose Hybrid Progressive Initialization in Sec. \ref{4.3} and Material-Based Optimization in Sec. \ref{4.4}.

\subsection{Physically Grounded Ellipsoid Representation}
\label{4.1}
\noindent\textbf{X-ray Physical Field.} 
X-ray imaging quantifies the cumulative attenuation of X-rays as they traverse an object, governed by the material-dependent absorption properties of the medium.
To formally describe this process, we define the X-ray Physical Field, where each spatial position $\mathbf{x} \in \mathbb{R}^3$ is characterized by the local energy absorption rate $\sigma(\mathbf{x}) \in \mathbb{R}^+$ of X-ray \cite{kak2001principles, hounsfield1973computerized, beutel2000handbook}.
When X-rays with an initial intensity $I_0$ propagate through the X-ray Physical Field, their energy is progressively attenuated by the materials they traverse.
Ultimately, the remaining intensity forms a projection image $I \in \mathbb{R}^{H \times W}$.
Mathematically, we represent an X-ray path as $\mathbf{r}(t) = \mathbf{o} + t \mathbf{d} \in \mathbb{R}^3$,
where $\mathbf{o}$ denotes the X-ray source position, $\mathbf{d}$ is the unit view direction vector, and $t$ varies between the entry $t_0$ and exit $t_n$ points of the object. According to the Beer-Lambert law \cite{swinehart1962beer}, the X-ray intensity $I'(\mathbf{r})$ after attenuation is given by:
\begin{equation}
I'(\mathbf{r}) = I_0 \exp\left( - \int_{t_0}^{t_n} \sigma(\mathbf{r}(t)) \, dt \right).
\label{eq:beer_lambert}
\end{equation}
In practice, the raw data is typically operated in logarithmic space to improve computational efficiency~\cite{kak2001principles}:
\begin{equation}
I(\mathbf{r}) = \log I_0 - \log I'(\mathbf{r}) = \int_{t_0}^{t_n} \sigma(\mathbf{r}(t)) \, dt.
\label{eq:I_0}
\end{equation}
Thus, each pixel intensity $I(\mathbf{r})$ in the projection image aggregates the material absorption along the X-ray path, providing the foundation for reconstructing internal structures.

\noindent\textbf{Material-Adaptive Ellipsoids.}
Inspired by the infinitesimal method \cite{courant1965introduction}, we represent the spatial distribution of materials using a set of $N$ ellipsoids $\{\mathbf{E}_i\}$.
Each ellipsoid \(\mathbf{E}_i \) encodes material properties and contains a center position $\mathbf{p_c}$ and covariance matrix \(\mathbf{\Sigma}_{\text{3D}}\).
Faithfully adhering to the X-ray Physical Field, we further assign a distinct, non-negative attenuation coefficient $\sigma_i$ to characterize the local absorption behavior.
Given an X-ray path $\mathbf{r}$ that traverses multiple ellipsoids, the corresponding pixel intensity is determined by the accumulated attenuation along the path:
\begin{equation}
\label{eq:seperate}
\begin{aligned}
I(\mathbf{r}) &= \int_{t_0}^{t_n} \sigma(\mathbf{r}(t)) \, dt \\
              &= \int_{t_0}^{t_1} \sigma(\mathbf{r}(t)) \, dt + \cdots + \int_{t_{n-1}}^{t_n} \sigma(\mathbf{r}(t)) \, dt \\
              &= \sigma_0 \int_{t_0}^{t_1} \, dt + \cdots + \sigma_{n-1} \int_{t_{n-1}}^{t_n} \, dt \\
              &= \sigma_0 l_0 + \sigma_1 l_1 + \cdots + \sigma_{n-1} l_{n-1},
\end{aligned}
\end{equation}
where $l_i=t_{i+1} - t_{i} \geq 0$ represents the segment length of the X-ray path within ellipsoid \(\mathbf{E}_i\).
Therefore, accurately measuring the accumulated attenuation requires precise determination of each segment length \( l_i \). In the following, we derive an explicit formulation for \( l_i \).

\noindent\textbf{Explicit Form of Segment Lengths.}
Our objective is to compute the segment length \( l_i \) for ellipsoid \(\mathbf{E}_i\) along the view direction \(\mathbf{d}\), a process illustrated in Figure \ref{fig:pipeline}(b).
To achieve this, we first transform \(\mathbf{E}_i\) from world to camera coordinates and then into Normalized Device Coordinates (NDC) space, aligning X-rays with the coordinate axes for computational simplicity.
To establish a reference for general segment length computation, we define the maximum segment length $l_{\text{max}}$ as a pivot, which corresponds to the ray path passing through the center of \(\mathbf{E}_i\):
\begin{equation}
\label{eq:lmax}
l_{\text{max}} = \frac{2}{\sqrt{\mathbf{d}^\top \mathbf{\Sigma}_{\text{3D}}^{-1} \mathbf{d}}}.
\end{equation}
Given $l_{\text{max}}$, we then compute the segment length \( l_i \) for an arbitrary point $\mathbf{u}$ inside the projected ellipse of \(\mathbf{E}_i\) in NDC space. The relationship is formulated as:
\begin{equation}
\label{eq:li}
l_i = l_{\text{max}} \times \sqrt{1 - \left( \frac{C - B^2}{A} \right)}, \text{ where}
\end{equation}
\begin{equation}
A = \mathbf{d}^\top \mathbf{\Sigma}_{\text{3D}}^{-1} \mathbf{d}, \quad B = \mathbf{a}^\top \mathbf{\Sigma}_{\text{3D}}^{-1} \mathbf{d}, \quad C = \mathbf{a}^\top \mathbf{\Sigma}_{\text{3D}}^{-1} \mathbf{a}.
\end{equation}
Here, \( \mathbf{a} = \mathbf{u} - \mathbf{p_c} \) is the displacement vector from the center of the ellipsoid \(\mathbf{p_c}\) to the point \(\mathbf{u}\).
The full derivation of \( l_{\text{max}} \) and \( l_i \) is provided in the supplementary materials.

\noindent\textbf{Segment Lengths with Intersections.}
As ellipsoids adjust their positions and shapes during optimization, intersections may occur, leading to ambiguities in the attenuation rates within overlapping regions.
To ensure that each spatial location corresponds to a single dominant material, resolving these complex overlaps is crucial to maintaining a physically consistent representation.
For optimization stability, we adopt a simple yet effective first-pass precedence strategy.
Given two intersecting ellipsoids along the X-ray path, we first sort them in ascending depth order, denoted by \(\mathbf{E}_i\) and \(\mathbf{E}_{i+1}\), where \(\mathbf{E}_i\) appears first along the ray.
The overlapping region $e_i$ is then exclusively assigned to $\mathbf{E}_i$ by adjusting the effective region of $\mathbf{E}_{i+1}$ to \( \mathbf{E}_{i+1} - e_i \).
This modification directly impacts the segment length computation, requiring additional updates to ensure consistency in attenuation accumulation, as shown in Figure \ref{fig:pipeline}(b).
To systematically handle different intersection scenarios, we introduce an efficient correction function $f$, as outlined in Algorithm \ref{alg:ellipsoid_lengths}. The final accumulated attenuation along the X-ray path is then expressed as:
\begin{equation}
I(\mathbf{r}) = \sum_{i=0}^{n-1} \sigma_i \cdot f \left( \frac{2}{\sqrt{A}} \times \sqrt{1 - \left( \frac{C - B^2}{A} \right)} \right),
\end{equation}
\begin{algorithm}[t]
\caption{Segment Length Correction with Intersections}
\label{alg:ellipsoid_lengths}
\begin{algorithmic}[1]
\REQUIRE $(z_0, z_1, \ldots, z_{n-1})$: Sorted depths of ellipsoids\\
$(l_0, l_1, \ldots, l_{n-1})$: Initial segment lengths without considering intersections
\ENSURE Updated segment lengths $\tilde{l}_0, \tilde{l}_1, \ldots, \tilde{l}_{n-1}$ and corrected effective regions

\FOR{$i = 0$ \TO $n-1$}
    \IF{$i == 0$}
        \STATE \textcolor{blue}{ $\tilde{l}_0 \gets l_0$}
        \STATE $z \gets z_0$, $l \gets l_0$
    \ELSE
        \IF{$z_i < z + \frac{1}{2}l$}
            \STATE \textcolor{red}{$\tilde{l}_i \gets \max(0, (z_i + \frac{1}{2}l_i) - (z + \frac{1}{2}l))$}
        \ELSE
            \STATE \textcolor{red}{$\tilde{l}_i \gets \min(l_i, (\frac{1}{2}l_i + z_i) - (z + \frac{1}{2}l))$}
        \ENDIF
        \IF{$\tilde{l}_i \neq 0$}
            \STATE Update the valid region of ellipsoid $\mathbf{E}_i$ as $[z_i + \frac{1}{2}l_i - \tilde{l}_i, z_i + \frac{1}{2}l_i]$
            \STATE $z \gets z_i$, $l \gets l_i$
        \ENDIF
    \ENDIF
\ENDFOR
\end{algorithmic}
\end{algorithm}

\begin{figure}[t]
    \centering
    \includegraphics[width=\linewidth]{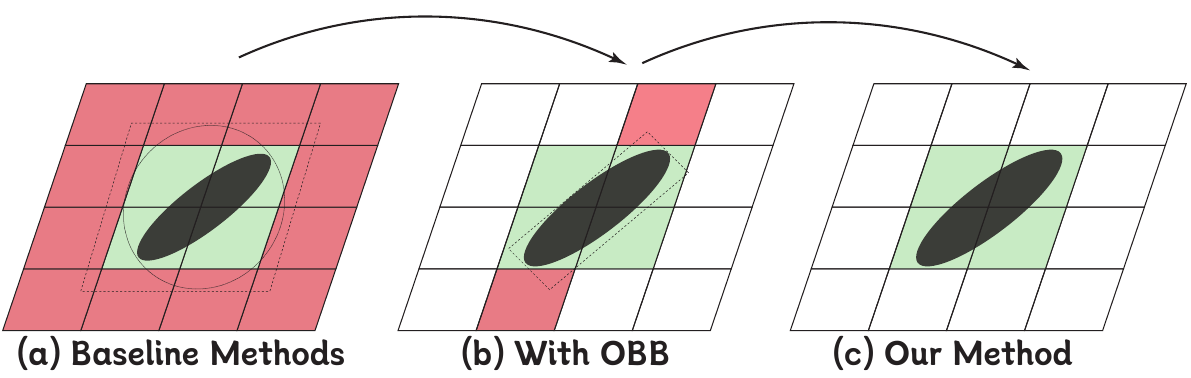}
    \captionsetup{type=figure}
    \vspace{-2em}
    \caption{\textbf{Pixel-Ellipse Association.} (a) The baseline, based on AABB \cite{toth1985aabb}, results in \textcolor{myred}{incorrect associations}. (b) The pixels selected by OBB~\cite{gottschalk1996obb}. (c) Our method only retains \textcolor{mygreen}{pixels} truly aligned with the projected ellipse to ensure physical faithfulness.}
    \vspace{-1em}
    \label{fig:overlap}
\end{figure}

\noindent\textbf{Physically Faithful Overlap Filtering.}
While resolving segment length inconsistencies ensures a physically valid attenuation modeling along the X-ray path, accurately associating ellipsoids with their corresponding pixels is equally critical for maintaining consistency in the projected space. Current methods often rely on coarse bounding-box approximations, leading to unintended pixel assignments that compromise reconstruction fidelity.
To address this, we introduce a refined Oriented Bounding Box (OBB) strategy that establishes precise pixel-ellipsoid associations. As illustrated in Figure \ref{fig:overlap}, existing approaches~\cite{3dgs, x_gaussian, r2_gaussian} typically rely on Axis-Aligned Bounding Boxes (AABB)~\cite{toth1985aabb}, which approximate the projected ellipse using a circumscribed circle. However, this method includes extraneous pixels (\textcolor{myred}{red region}) that do not truly overlap with the projected ellipse, leading to physically inconsistent assignments.
In contrast, we first determine the OBB via eigenvalue decomposition, aligning it with the principal axes of the projected ellipse.
Then, we filter out irrelevant pixels while preserving only the valid tiles (\textcolor{mygreen}{green region}) that are closely related to the actual projected ellipse. By eliminating spurious pixel association, our method enhances the physical consistency and numerical stability of the reconstruction process.

\subsection{Hybrid Progressive Initialization}
\label{4.3}
It is widely acknowledged that initializing point clouds with detailed geometric information accelerates model convergence~\cite{3dgs, x_gaussian, huang20242d}.
However, conventional initialization techniques fail in the context of X-ray imaging.
Specifically, SfM-based methods~\cite{schoenberger2016sfm, 3dgs} estimate point clouds by feature matching across multiple images. While in X-ray imaging, variations in ray paths significantly alter the remaining energy intensity, leading to feature mismatches and unreliable correspondences.
Similarly, point cloud regression methods such as Dust3R~\cite{wang2024dust3rgeometric3dvision, zhang2024monst3r, yang2024gaussianobject} directly infer 3D structures from images but fail in X-ray reconstruction due to domain shifts, as they are trained on natural images and do not account for X-ray-specific attenuation properties.

\begin{figure}[t]
    \centering
    \includegraphics[width=\linewidth]{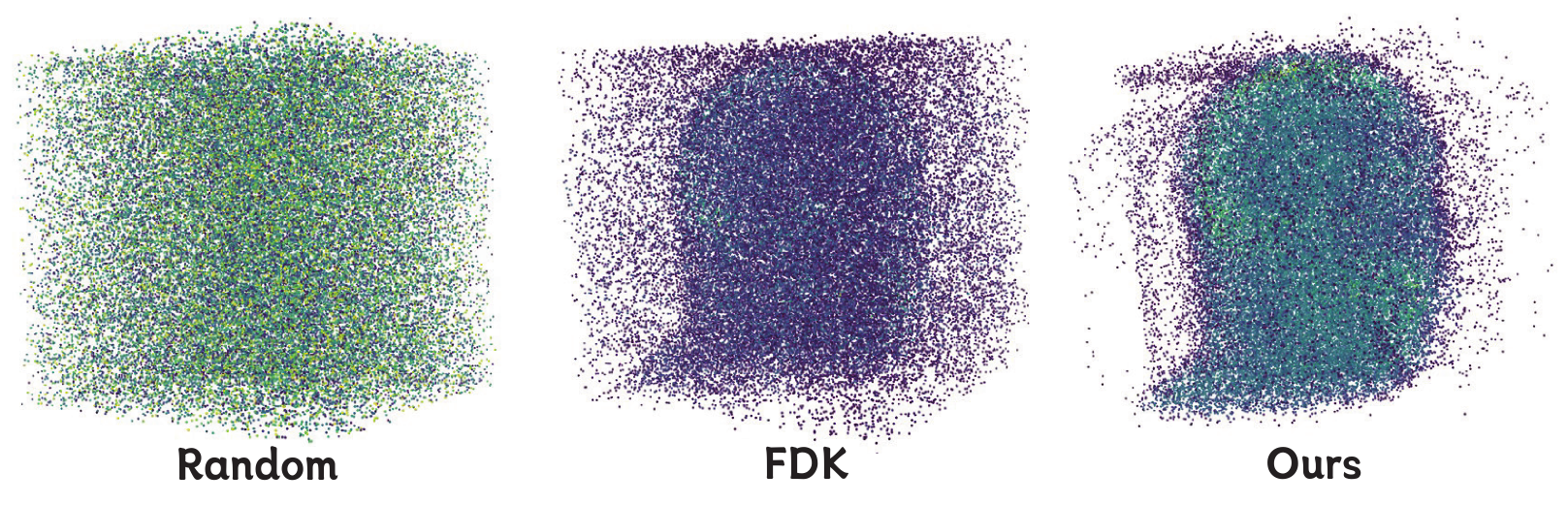}
    \captionsetup{type=figure}
    \vspace{-1.6em}
    \caption{\textbf{Comparison of Initialization Methods.} Our Hybrid Progressive Initialization produces clear shapes, whereas Random~\cite{x_gaussian} and FDK~\cite{r2_gaussian} initialization sacrifice the geometry prior.}
    \vspace{-1.3em}
    \label{fig:initialization_main}
    \vspace{-4pt}
\end{figure}

To address this challenge, we design a Hybrid Initialization strategy that progressively refines the geometric structure through a sequence of iterative methods. We compare the results in Figure \ref{fig:initialization_main}.
Our process begins with Conjugate Gradient Least Squares (CGLS) \cite{hestenes1952methods, bjorck1979conjugate, kak2001principles}, which efficiently provides a coarse global estimate.
Next, we refine this estimate using Simultaneous Algebraic Reconstruction Technique (SART) \cite{andersen1984simultaneous}, leveraging its capability to enhance local details and correct early-stage inconsistencies.
Finally, we incorporate Total Variation (TV) regularization \cite{rudin1992nonlinear} into the initialization process to suppress noise and artifacts, preserving critical structural edges.
For each 3D point obtained from our initialization, we define an ellipsoid centered at that position, assigning it a random covariance matrix \(\mathbf{\Sigma}_{\text{3D}}\) and an attenuation coefficient \(\sigma_i\).
We subsequently devise an optimization method for these parameters for better modeling the detailed material distribution.

\subsection{Material-Based Optimization}
\label{4.4}
\begin{figure}[t]
    \centering
    \includegraphics[width=\linewidth]{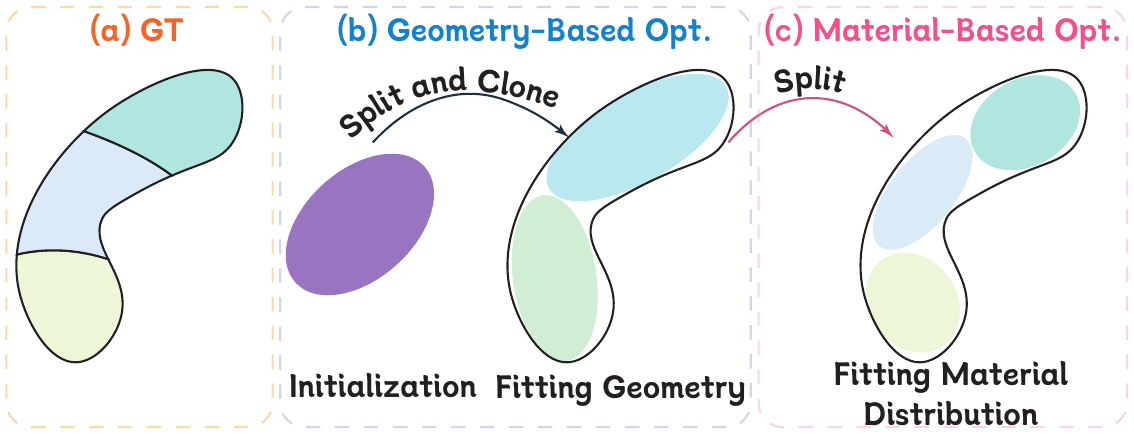}
    \captionsetup{type=figure}
    \vspace{-18pt}
    \caption{\textbf{Illustration of Adaptive Optimization Strategy.} \textbf{\textcolor{Orange}{(a) Ground-truth Geometry and Material Distribution}}, with different colors indicating distinct materials. \textbf{\textcolor{DeepBlue}{(b) Geometry-Based Optimization}} \cite{3dgs}, which fits the ellipsoids closely to the object geometry. \textbf{\textcolor{MyPink}{(c) Our Material-Based Optimization}}, which further refines the ellipsoids to capture the material distribution.}
    \vspace{-16pt}
    \label{fig:opt}
\end{figure}
Mainstream optimization strategy \cite{3dgs, x_gaussian, r2_gaussian, huang20242d} involves splitting and cloning in regions with poor structural fidelity while pruning those with negligible opacity.
However, this geometry-based optimization fails to capture the true material composition within objects.
As depicted in Figure \ref{fig:opt}(a), the ground-truth object consists of three distinct materials, represented by different colors.
While the geometry-based optimization strategy successfully reconstructs the overall shape, it captures only two intermediate material types instead of all three, as shown in Figure \ref{fig:opt}(b). This highlights the limitations of purely geometry-centric approaches in faithfully modeling heterogeneous material compositions.

The challenge arises from the continuous fluctuation of an ellipsoid’s attenuation coefficient, which hinders stable convergence and prevents accurate material differentiation. To address this, we propose a material-based optimization strategy that explicitly accounts for the heterogeneous material distributions inherent in the X-ray Physical Field.
Our approach is designed to identify material boundaries through local density estimation. Concretely, we randomly sample a subset of ellipsoids and compute the k-nearest neighbors \cite{peterson2009k}, averaging their distances and analyzing the attenuation gradient.
Regions with high density and steep gradients indicate material transitions, necessitating a finer adjustment. Therefore, to enhance fidelity, we selectively split ellipsoids in these complex regions, scaling each by a factor of 1.6, which follows empirical heuristics in prior studies~\cite{3dgs, x_gaussian, r2_gaussian}.
As presented in Figure \ref{fig:opt}(c), our strategy significantly improves material separation, faithfully capturing the internal composition of objects.

\begin{table*}
\centering
\scriptsize
\adjustbox{width=1.0\linewidth}{
\begin{tabular}{r|ccc|ccc|ccc|ccc}
    \Xhline{1.5pt}
    \rowcolor[HTML]{C0C0C0} 
    {} &
    \multicolumn{3}{c|}{Human Organ 10-views~\cite{r2_gaussian}} &
    \multicolumn{3}{c|}{Daily Object 10-views~\cite{r2_gaussian}} &
    \multicolumn{3}{c|}{Human Organ 5-views~\cite{r2_gaussian}} &
    \multicolumn{3}{c}{Daily Object 5-views~\cite{r2_gaussian}} \\ 
    
    \rowcolor[HTML]{C0C0C0} 
    {Method} & 
    PSNR$\uparrow$ & SSIM$\uparrow$ & LPIPS*$\downarrow$ &
    PSNR$\uparrow$ & SSIM$\uparrow$ & LPIPS*$\downarrow$ &
    PSNR$\uparrow$ & SSIM$\uparrow$ & LPIPS*$\downarrow$ &
    PSNR$\uparrow$ & SSIM$\uparrow$ & LPIPS*$\downarrow$ \\ 
    \Xhline{1.5pt}
    
    \multicolumn{13}{c}{\textit{Traditional Methods}} 
    \\ \hline
    
    FDK ~\cite{feldkamp1984practical} &
    12.35 & 0.675 & 291.2 &
    16.52 & 0.716 & 259.1 &
    8.15 & 0.618 & 310.6 &
    14.42 & 0.688 & 283.7 \\
    
    SART ~\cite{andersen1984simultaneous} &
    13.23 & 0.691 & 284.8 &
    17.69 & 0.724 & 247.3 &
    9.31 & 0.634 & 303.4 &
    15.68 & 0.663 & 293.5 \\
    
    \hline
    \multicolumn{13}{c}{\textit{Deep Learning-based Methods}} \\
    \hline

    TensoRF ~\cite{chen2022tensorf} &
    16.61 & 0.928 & 182.5 &
    24.19 & 0.946 & 153.4 &
    12.32 & 0.895 & 189.6 &
    18.27 & 0.922 & 210.8 \\

    NeAT ~\cite{ruckert2022neat} &
    16.22 & 0.934 & 185.3 &
    25.15 & 0.957 & 155.2 &
    11.08 & 0.887 & 188.3 &
    17.29 & 0.908 & 211.3 \\

    NAF ~\cite{zha2022naf} &
    17.89 & 0.925 & 193.2 &
    25.44 & 0.949 & 151.9 &
    11.19 & 0.894 & 197.1 &
    17.02 & 0.923 & 208.5 \\
    
    SAX-NeRF ~\cite{sax_nerf} &
    19.32 & 0.945 & 186.4 &
    25.38 & 0.979 & 143.6 &
    14.18 & 0.901 & 191.2 &
    19.09 & 0.948 & 204.9 \\
    
    X-Gaussian ~\cite{x_gaussian} &
    22.88 & 0.947 & 130.3 & 
    22.91 & 0.982 & 79.12 &
    17.23 & 0.947 & 176.4 & 
    20.31 & 0.961 & 108.1 \\
    
    R$^2$-Gaussian ~\cite{r2_gaussian} &
    \secondcolor 33.72 & \secondcolor 0.967 & \secondcolor 85.97 &
    \secondcolor 41.93 & \bestcolor 0.986 & \secondcolor 54.31 &
    \secondcolor 31.12 & \secondcolor 0.956 & \secondcolor 109.7 &
    \secondcolor 34.52 & \secondcolor 0.965 & \secondcolor 82.46 \\
    
    \textbf{Ours} &
    \bestcolor 35.71 & \bestcolor 0.980 & \bestcolor 71.03 &
    \bestcolor 42.80 & \secondcolor 0.983 & \bestcolor 45.64 &
    \bestcolor 32.34 & \bestcolor 0.963 & \bestcolor 103.2 &
    \bestcolor 37.41 & \bestcolor 0.970 & \bestcolor 81.02 \\
    
    \Xhline{1.5pt}
\end{tabular}
}
\vspace{-1.2em}
\caption{\textbf{Results of Quantitative Comparison (\S~\ref{sec:Comparison}).} We compare our X-Field with: 
    (a) Traditional X-ray reconstruction method: FDK\cite{feldkamp1984practical}, SART\cite{andersen1984simultaneous}. 
    (b) Deep Learning-based methods: TensoRF \cite{chen2022tensorf}, NeAT \cite{ruckert2022neat}, NAF \cite{zha2022naf}, SAX-NeRF \cite{sax_nerf}, X-Gaussian \cite{x_gaussian}, and R$^2$-Gaussian \cite{r2_gaussian}. 
    We report LPIPS* = LPIPS $\times 10^3$. 
    We mark out \colorbox{yzybest}{best} and \colorbox{yzysecond}{second best} method for all metrics.}
\label{tab:all_comparison}
\vspace{-16pt}
\end{table*}

\label{sec:experiment}
\section{Experiment}

\subsection{Experiment Settings}
\label{sec:sup_base_implementation}
\noindent\textbf{Dataset.}  Following the literature convention \cite{sax_nerf, x_gaussian},
we conduct experiments on a large-scale X3D dataset containing 15 scenes with two collections: Human Organs, derived from real-world medical datasets, to evaluate model performance in the medical domain; and Daily Objects, generated from synthetic datasets, to assess generalization ability. Specifically, chest scans are sourced from LIDC-IDRI~\cite{armato2011lung}, pancreas CT scans from Pancreas-CT~\cite{roth2016pancreasct}. The remaining objects from are obtained from VOLVIS~\cite{Philips2022} and the open scientific visualization dataset~\cite{klacansky2017scivis}.
We adopt the tomography toolbox TIGRE \cite{biguri2016tigre} to capture projections from CT volumes in the range of \(0^\circ \sim 180^\circ\) with minor scatter and electric noise. For highly sparse-view novel view synthesis, 5 and 10 views are used for training and 50 samples are used for testing. To further assess model performance and scalability, we generate 50, 25, and 15 views for evaluating the performance under sparse-view synthesis.


\noindent\textbf{Baselines.} We compare X-Field with state-of-the-art 3D X-ray reconstruction methods, including TensoRF \cite{chen2022tensorf}, NeAT \cite{ruckert2022neat}, NAF \cite{zha2022naf}, SAX-NeRF \cite{sax_nerf}, X-Gaussian \cite{x_gaussian}, and R$^2$-Gaussian \cite{r2_gaussian}. TensoRF, NeAT, NAF, and SAX-NeRF are NeRF-based methods designed for efficient reconstruction, with SAX-NeRF achieving SOTA performance among them by incorporating a transformer architecture as the model backbone. X-Gaussian and R$^2$-Gaussian are 3DGS-based methods, where X-Gaussian focuses on novel view synthesis, and R$^2$-Gaussian  extends its applicability for CT reconstruction by introducing voxelization. We also compare against traditional CT reconstruction methods including FDK \cite{feldkamp1984practical} and SART \cite{andersen1984simultaneous}. The novel view images are obtained by leveraging TIGRE for rendering.

\begin{table}[t]
\centering
\scriptsize
\adjustbox{width=1\linewidth}{
    \begin{tabular}{r|cc|cc}
    \Xhline{1.5pt}
    \rowcolor[HTML]{C0C0C0}
    {\#Views} & \multicolumn{2}{c|}{\textbf{5}} & \multicolumn{2}{c}{\textbf{10}} \\ 
    
    \rowcolor[HTML]{C0C0C0} 
    {Method} & PSNR $\uparrow$ & SSIM $\uparrow$ & PSNR $\uparrow$ & SSIM $\uparrow$ \\ 
    \Xhline{1.5pt} 
    FDK & 15.20 & 0.144 & 19.95 & 0.257 \\
    ASD\_POCS & 24.89 & 0.731 & 27.50 & 0.787 \\
    X-Gaussian & 17.57 & 0.688 & 18.44 & 0.537 \\
    R$^2$-Gaussian & 26.83 & 0.804 & 29.28 & \cellcolor[HTML]{FFCCC9} 0.946 \\
    \textbf{Ours} & \cellcolor[HTML]{FFCCC9} 28.04 & \cellcolor[HTML]{FFCCC9} 0.815 & \cellcolor[HTML]{FFCCC9} 33.26 & 0.911 \\
    \Xhline{1.5pt}
    \end{tabular}
}
\vspace{-1.2em}
\caption{\textbf{CT Reconstruction Comparison}(\S~\ref{sec:Comparison}). Ours achieve comparable results across different methods under 5 and 10 views.}
\label{tab:ablation_ct}
\vspace{-16pt}
\end{table}

\noindent\textbf{Metrics.} We adopt peak signal-to-noise ratio (PSNR) \cite{psnr} to assess the quality of rendered images, structural similarity index measure (SSIM) \cite{ssim} to measure consistency between predicted images and ground-truths, and Learned Perceptual Image Patch Similarity (LPIPS) \cite{zhang2018perceptual} to analyze the perceptual quality in high-level feature space. For clarity, we report LPIPS as LPIPS* = LPIPS \( \times 10 ^ 3\) instead.

\noindent\textbf{Implementation Details.} For X-ray novel view synthesis, we evaluate learning-based methods using their official implementations. While traditional methods like FDK and SART were originally designed for CT reconstruction, we first reconstruct CT volumes from sparse inputs and then use TIGRE to render novel X-ray projections at specified camera poses, comparing them with the ground truth.
For CT reconstruction, we follow \cite{ruckert2022neat, sax_nerf, x_gaussian} to synthesize multiple novel-view X-ray images from sparse inputs and reconstruct CT volumes using a total of 100 images. Specifically, geometry-based optimization begins after 10,000 epochs, and material-based optimization starts after 15,000 epochs. 

\begin{table}
    \centering
    \scriptsize
\adjustbox{width=1.0\linewidth}{
    \begin{tabular}{r|ccc}
    \Xhline{1.5pt}
    \rowcolor[HTML]{C0C0C0} 
    Method      & PSNR $\uparrow$  & SSIM $\uparrow$ & LPIPS* $\downarrow $\\ \hline
    w/o Material Opt. & 34.78 & 0.941 & 73.45  \\
    w/o Overlap Filter & 34.59 & 0.937 & 74.32 \\ 
    w/o Intersection   &  33.84 &  0.929 &  76.60  \\
    w/o Segment Length & 27.48 & 0.875 & 87.83 \\ 
    \textbf{Ours} & \bestcolor35.03 & \bestcolor0.953 & \bestcolor72.12 \\ 
    \Xhline{2\arrayrulewidth}
    \end{tabular}
}
\vspace{-1.2em}
    \caption{\textbf{Proposed Components (\S~\ref{sec:ablation}).} Our method achieves the best performance when all components are included.} 
    \label{tab:ablation_component}
\vspace{-16pt}
\end{table}

\begin{figure*}[htb]
    \centering
    \includegraphics[width=\linewidth]{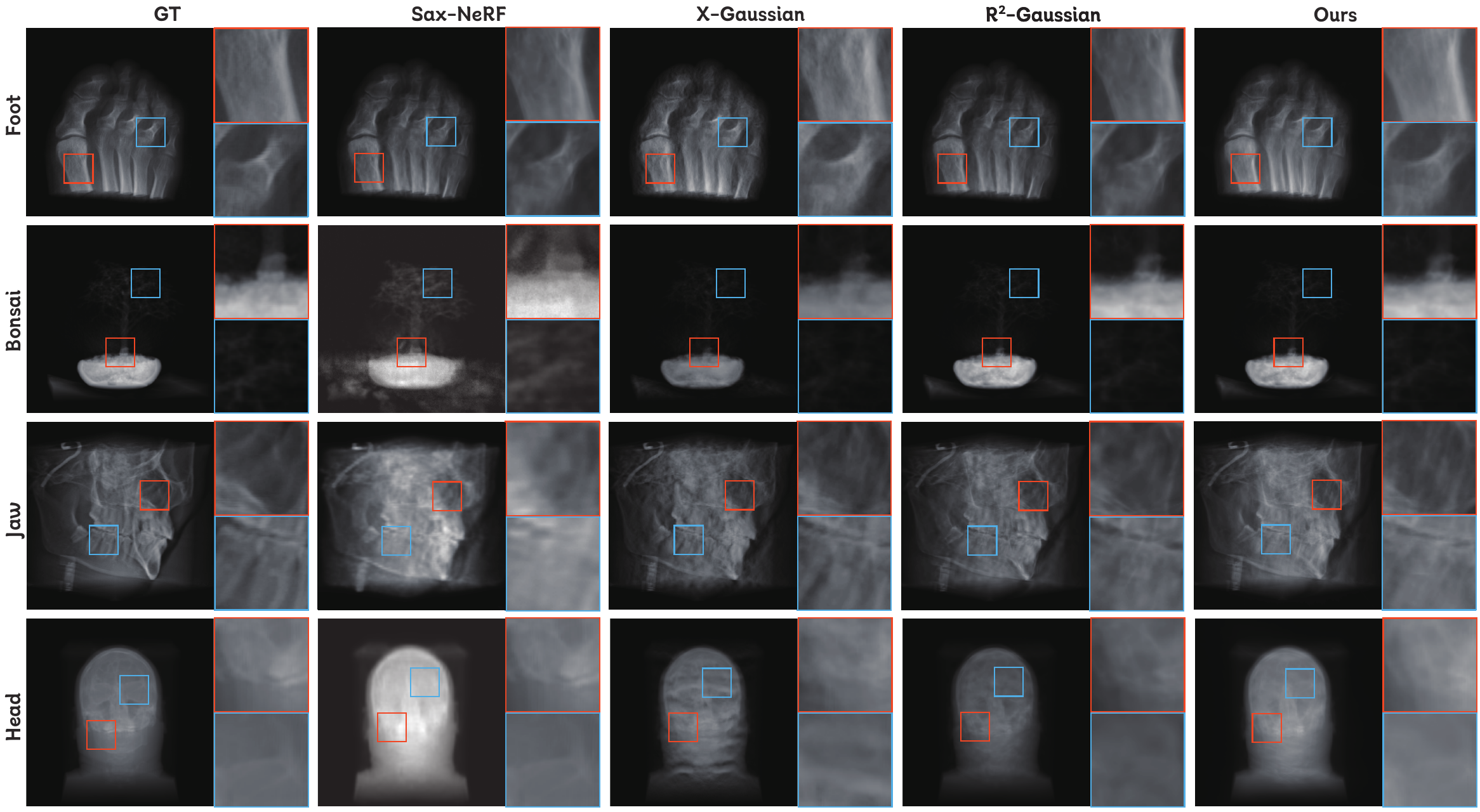}
    \captionsetup{type=figure}
    \vspace{-2.0em} 
    \caption{\textbf{Qualitative Comparison of NVS (\S~\ref{sec:Comparison}).} We present visual examples of reconstructed images across four cases trained with 10 views. Our results demonstrate superior visual quality, richer details, and fewer spatial artifacts. Please \faSearch ~zoom in for a closer examination.}
    \vspace{-0.6em} 
    \label{fig:compare}
    \vspace{-12pt}
\end{figure*}

\subsection{Comparison with State-of-the-Art Methods}
\label{sec:Comparison}
We present the quantitative results of X-ray NVS and CT reconstruction, discussing the qualitative results across all scenes, highlighting the superior performance of X-Field.

\noindent\textbf{Discussion on NVS Quantitative Results.}
We compare X-Field with two traditional methods (FDK, SART), three NeRF-based methods (TensoRF, NeAT, NAF), and three SOTA methods (SAX-NeRF, X-Gaussian, and R$^2$-Gaussian). Table \ref{tab:all_comparison} reports the quantitative results of highly sparse views (10 views and 5 views) X-ray NVS. Note that we report quantitative results as the mean results of scenes under the same setting, and scene-wise results are presented in the supplementary material. X-Field demonstrates superior performance in reconstructing X-ray novel views across most scenarios, surpassing the state-of-the-art R$^2$-Gaussian in all metrics. 
Specifically, in the human organ reconstruction setting, X-Field consistently outperforms the state-of-the-art R$^2$-Gaussian, achieving higher SSIM scores and competitive PSNR and LPIPS values,  highlighting its effectiveness in complex reconstruction scenarios. 

\begin{table}
    \centering
    \scriptsize
\adjustbox{width=0.9\linewidth}{
    \begin{tabular}{r|ccc}
    \Xhline{1.5pt}
    \rowcolor[HTML]{C0C0C0} 
    Initialization      & PSNR $\uparrow$  & SSIM $\uparrow$ & LPIPS* $\downarrow $\\ \hline
    Random & 37.85 & 0.966 & 60.02  \\
    FDK\cite{feldkamp1984practical} & 37.96 & 0.967 & \bestcolor59.81 \\ 
    \textbf{Ours} & \bestcolor38.67 & \bestcolor0.969 & 59.95 \\ 
    \Xhline{2\arrayrulewidth}
    \end{tabular}
}
\vspace{-1.2em}
    \caption{\textbf{Initialization Methods (\S~\ref{sec:ablation}).} Our initialization with detailed geometry information improves reconstruction accuracy. }
    \label{tab:ablation_initialization}
\vspace{-20pt}
\end{table}

\noindent\textbf{Discussion on NVS Qualitative Results.}
Figure \ref{fig:compare} presents visual comparisons between X-Field and multiple methods, including SAX-NeRF, X-Gaussian, and R$^2$-Gaussian. These highly sparse view settings provide limited information, resulting in artifacts of varying severity across all methods. SAX-NeRF reconstructs the overall structure but introduces noticeable blurry artifacts, particularly in the bonsai scene. X-Gaussian produces line and wave-pattern artifacts, which are prominent in the head scene. While R$^2$-Gaussian  performs better than the other baselines, it exhibits flaws in the fine details of the bone structure. For example, R$^2$-Gaussian 
 introduces black linear artifacts in the bone region in the foot scene, where ours produces smoother textures. In summary, our method is able to effectively mitigate blurry, line, and wave-pattern artifacts while maintaining smoothness in distinct areas and on object surfaces. More visual comparisons are in supplementary video.

\begin{figure*}[t]
    \centering
    \includegraphics[width=\linewidth]{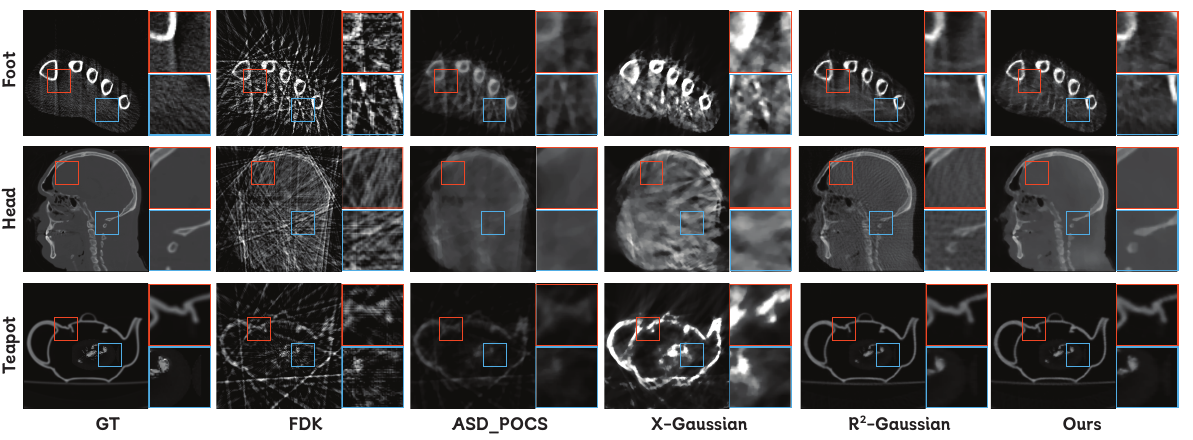}
    \captionsetup{type=figure}
    \vspace{-2.0em} 
    \caption{\textbf{Qualitative Comparison of CT Reconstruction(\S~\ref{sec:Comparison}).} Our method produces clearer textures, more refined anatomical structures, and fewer artifacts, particularly in high-contrast regions such as the cranial cavity. Please \faSearch ~zoom in for a closer examination.}
    \vspace{-0.9em} 
    \label{fig:ct}
    \vspace{-8pt}
\end{figure*}

\noindent\textbf{Discussion on CT Reconstruction Results.}
We compare X-Field with two traditional learning-free algorithms, FDK~\cite{feldkamp1984practical} and ASD\_POCS~\cite{asd}, as well as two learning-based methods including X-Gaussian~\cite{x_gaussian} and R$^2$-Gaussian~\cite{r2_gaussian} for sparse-view CT reconstruction. Specifically, we assess the performance of 3DGS-based methods and X-Field by generating novel view images from sparse input projections (5 and 10 views) and reconstructing CT scans using ASD\_POCS with a total of 100 views. The quantitative results, presented in Table~\ref{tab:ablation_ct}, demonstrate that X-Field consistently achieves the best performance across all scenarios.
In particular, when generating novel X-ray projections from 10 input views, X-Field combined with ASD\_POCS achieves a PSNR of 33.26 and SSIM of 0.911, a comparable result with the SOTA method, R$^2$-Gaussian.

Figure~\ref{fig:ct} shows qualitative results of sparse-view CT reconstruction of foot, head, and teapot scans using 10 input views. Without novel view projections, FDK introduces streak artifacts, while ASD\_POCS results in blurred structural details. The NVS of X-Gaussian produces clearer reconstructions but introduces shadow artifacts. In contrast, our method and R$^2$-Gaussian achieve superior visual quality, with X-Field delivering better results by recovering smoother textures and reducing needle-like artifacts, especially in delicate structures such as the cranial region.

\begin{table}
    \centering
    \tiny
\adjustbox{width=0.9\linewidth}{
    \begin{tabular}{c|ccc}
    \Xhline{1.5pt}
    \rowcolor[HTML]{C0C0C0} 
    \#Views      & PSNR $\uparrow$  & SSIM $\uparrow$ & LPIPS* $\downarrow $\\ \hline
    5 & 31.61 & 0.933 & 92.23  \\
    10   &  35.11 &  0.959 &  83.72  \\
    15 & 38.31 & 0.968 & 82.91 \\ 
    25 & 41.71 & 0.979 & 75.26 \\ 
    50 & \bestcolor42.61 & \bestcolor0.993 & \bestcolor 61.19 \\ 
    \Xhline{2\arrayrulewidth}
    \end{tabular}
}
\vspace{-1.2em}
    \caption{\textbf{Number of Input Views (\S~\ref{sec:ablation}).} Our method's performance improves as the number of input views increases.}
    \label{tab:ablation_view}
    \vspace{-20pt}
\end{table}

\subsection{Ablation Study}
\label{sec:ablation}
To comprehensively assess the performance of X-Field, we evaluate the impact of the proposed components, compare different initialization strategies, and evaluate X-Field under various input view settings from 5 to 50 views. 

\noindent\textbf{Component Analysis.}  
Table \ref{tab:ablation_component} evaluates the effects of individual components on reconstruction performance. We observe that Material-Based Optimization and Overlap Filter have a limited impact on reconstruction quality, focusing instead on improving rendering efficiency and aligning with X-ray physical properties. Removing the Intersection Module (Sec.~\ref{4.1}) leads to substantial performance drops, with a PSNR drop of 1.19 db, and an SSIM decrease of 0.024, underscoring its importance in preserving structural integrity.

Segment Length is a fundamental component of our ellipsoid representation, defined in Sec.~\ref{4.1} for capturing the distance each ray traverses within the ellipsoid. Removing it also severely impacts the model's ability to render novel views, leading to significant performance degradation, with a 7.6 drop in PSNR and an 11.71 increase in LPIPS. These results underscore the importance of Segment Length in enabling X-Field to achieve high-quality reconstructions.

\noindent\textbf{Initialization Analysis}  
We compare the our hybrid initialization with random initialization used in X-Gaussian and FDK \cite{feldkamp1984practical} introduced in R$^2$-Guassian.  
As shown in Figure~\ref{fig:initialization_main}, our method provides a point cloud with clearer boundaries and reduced noise, facilitating initialization.
Table \ref{tab:ablation_initialization} shows that both FDK and our hybrid initialization outperform random initialization. For random initialization did not provide geometry prior. While FDK achieves a slightly lower LPIPS, it also results in lower PSNR and SSIM. In contrast, ours improves both PSNR and SSIM, demonstrating its effectiveness in enhancing reconstruction quality.

\begin{figure}[t]
\vspace{-0em}
    \centering
    \includegraphics[width=\linewidth]{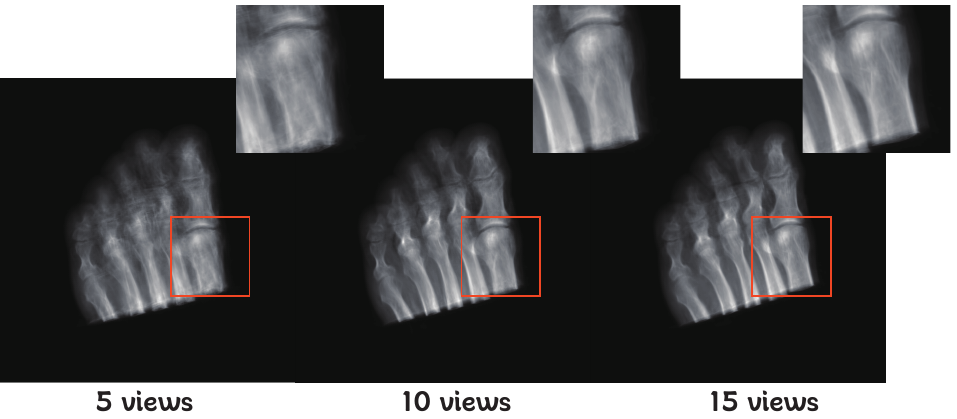}
    \captionsetup{type=figure}
    \vspace{-1.8em}
   \caption{\textbf{Input View Numbers (\S~\ref{sec:ablation}).} Five views enable good NVS, and additional input views enhance texture smoothness.}
    \vspace{-1.0em}
    \label{fig:ablation_view}
    \vspace{-8pt}
\end{figure}

\noindent\textbf{Input View Number Analysis.} 
To further demonstrate the scalability of X-Field, we conduct experiments to assess the effect of the input view number on reconstruction performance. As shown in Table \ref{tab:ablation_view}, as the number of input views increases, the performance is consistently enhanced. When using 50 views as input, X-Field achieves performance comparable to work designed for sparse view X-Ray reconstruction \cite{x_gaussian, sax_nerf}. We further show in Figure \ref{fig:ablation_view} that, with the increase in the input number, the synthesis X-ray images exhibit smoother and clearer bone textures.

\section{Conclusion}
We present X-Field, the first physically grounded representation derived from the penetration and attenuation properties of X-rays for X-ray imaging.
To model diverse materials within the internal structures of objects, we introduce material-adaptive ellipsoids with distinct attenuation coefficients.
To compute pixel intensity, we design a segment length algorithm that incorporates ellipsoid intersections, enabling accurate estimation of each material's X-ray energy absorption.
To improve reconstruction performance, we optimize the ellipsoids along material boundaries and refine the geometric properties during initialization.
Our proposed X-Field significantly outperforms state-of-the-art methods in both X-ray NVS and CT reconstruction, demonstrating strong potential for medical applications.
Furthermore, we provide insights into the fundamental design of X-ray imaging, which can be extended to other tasks such as the reconstruction of translucent objects.

{
    \small
    \bibliographystyle{ieeenat_fullname}
    \bibliography{main}
}

\end{document}